\documentclass[letterpaper, 10 pt, conference]{ieeeconf}      
\usepackage{amssymb,graphicx,amsmath,color,algorithm,algorithmic,url}
\usepackage[english]{babel}
\usepackage{blindtext}
\usepackage{afterpage}
\usepackage[noadjust]{cite} 
\usepackage{hhline}
\usepackage[draft]{hyperref}
\usepackage{diagbox} 
\usepackage{multirow}
\usepackage{arydshln}
\usepackage[table]{xcolor}
\usepackage{booktabs}
\usepackage{makecell}
\usepackage{array}
\usepackage{tabularx}

\newcolumntype{C}[1]{>{\centering\arraybackslash}m{#1}}
\newcolumntype{M}[1]{>{\centering\arraybackslash}m{#1}}
\newcommand{\lightrowrule}{%
    \arrayrulecolor{black!15}%
    \specialrule{0.3pt}{1.5pt}{1.5pt}%
    \arrayrulecolor{black}%
}

\IEEEoverridecommandlockouts        
\title{\LARGE\bfseries
Tactile In-Hand Object Rotation: A Comparative Study of Hand~Morphology in Simulation and Hardware
}
\author{
Wenhao He,
Max Yang, 
Chenghua Lu, 
Yijiong Lin,
David A.W. Barton,
Nathan F. Lepora,
 \thanks{All authors are with Department of Engineering Mathematics and Bristol Robotics Laboratory, University of Bristol, Bristol BS8 1UB, U.K.}
\thanks{$^{*}$Corresponding author:
   Nathan F. Lepora (n.lepora@bristol.ac.uk).}%
}

\begin{document}
\bstctlcite{IEEEexample:BSTcontrol}
\maketitle
\thispagestyle{empty}
\pagestyle{empty}
\begin{abstract}
Continuous in-hand rotation requires coordinated contact
maintenance and renewal, linking manipulation performance to
both hand geometry and tactile feedback. We investigate these
relationships through rotational progress, finger participation,
and fingertip loading on two versions of the low-cost LEAP Hand with
anthropomorphic and symmetric finger arrangements. Both hands are integrated with
TacTip vision-based tactile sensors and use a shared reinforcement learning
framework. Simulation evaluates full tactile, binary touch, and
proprioception-only observations on 12 unseen objects; hardware
experiments evaluate full tactile policies on ten objects under
multiple hand orientations. In simulation, full tactile
observations increase mean normalized rotation per
episode over proprioception alone by 94.8\% for the
anthropomorphic hand and 66.9\% for the symmetric hand, with
longer mean episode durations and lower mean total fingertip
loads. Hardware contact measurements reveal more differentiated
finger participation in the anthropomorphic hand, while
cumulative loading is more evenly distributed across the
symmetric hand's fingers. Overall, the symmetric hand achieves a 32.0\%
higher mean rotation count and a 25.8\% higher aggregate
rotation rate, together with longer manipulation duration.
Additional hardware force recordings show a 19.9\% lower
mean total fingertip load for this layout. These findings
relate finger arrangement to contact organization and load
sharing, and show that richer tactile information supports
greater task progress at lower mean loading in both
simulated embodiments.
\end{abstract}


\section{INTRODUCTION}\label{sec:Intro}
A robotic hand's geometry determines where its fingertips
can reach and how they approach an object. During sustained
rotation, these geometric constraints interact with the need
to support the object, permit motion, and renew contact under
different gravity directions. The forces carried by the
fingertips and the distribution of contact changes across
fingers characterize how these interactions develop during
manipulation. Learning-based controllers have enabled
continuous fingertip rotation using proprioceptive and tactile
observations~\cite{qi2023inhand,qi2023general,yang2024anyrotate}, providing an experimental basis for
examining these relationships across hand layouts.

The question considered here is how finger arrangement
changes both the organization of fingertip contact and loading
and the benefit obtained from tactile feedback. Similar task
scores can arise from different distributions of persistent
contact, contact-state changes, and accumulated fingertip load.
Examining these quantities together provides a basis for
interpreting how each layout performs the task. Comparing full
tactile, binary touch, and proprioception-only observations
then establishes how task progress, duration, and average
loading change as more local contact information becomes
available.

We study anthropomorphic and symmetric configurations of
the low-cost LEAP Hand~\cite{shaw2023leap}, each with 16 actuated joints and
4 integrated vision-based (TacTip) tactile fingertips~\cite{lepora2021soft}. We adopt the task and
reinforcement learning framework of AnyRotate~\cite{yang2024anyrotate}  as a shared experimental protocol. The comparison uses a common
tactile observation interface and learning procedure to examine
differences across finger arrangements and observation
conditions. Simulation evaluates the three observation
conditions on 12 unseen objects with randomized hand
orientations and rotation axes. Hardware experiments evaluate
full tactile policies on ten objects under multiple hand
orientations, with additional force recordings following the
corresponding evaluation settings.

The results associate the two layouts with distinct patterns
of finger participation and cumulative loading. On hardware,
the anthropomorphic thumb and middle finger maintain contact
more persistently and contribute a larger share of accumulated
load, with more frequent contact changes at the index and
little fingers. The symmetric hand distributes participation
more evenly, and its cumulative loads are more uniformly
shared in both simulation and hardware. In the simulation comparison, richer tactile observations improve mean normalized rotation and manipulation duration while reducing mean fingertip loading in both hands.

The main contributions of this work are:
\begin{enumerate}
\item We compare two hand morphologies under three observation
conditions, quantifying changes in normalized rotation,
manipulation duration, and mean fingertip loading on unseen
simulated objects.

\item We relate fingertip workspace arrangement to hardware
contact persistence and transitions, and to cumulative load
sharing in simulation and hardware, identifying differentiated
and more evenly distributed patterns of finger contribution.

\item We implement two low-cost LEAP--TacTip hand configurations
and evaluate their continuous fingertip rotation on ten physical
objects under multiple hand orientations, combining manipulation
outcomes with complementary contact and loading measurements.
\end{enumerate}

\begin{figure*}[t]
    \centering
    \includegraphics[width=\textwidth]{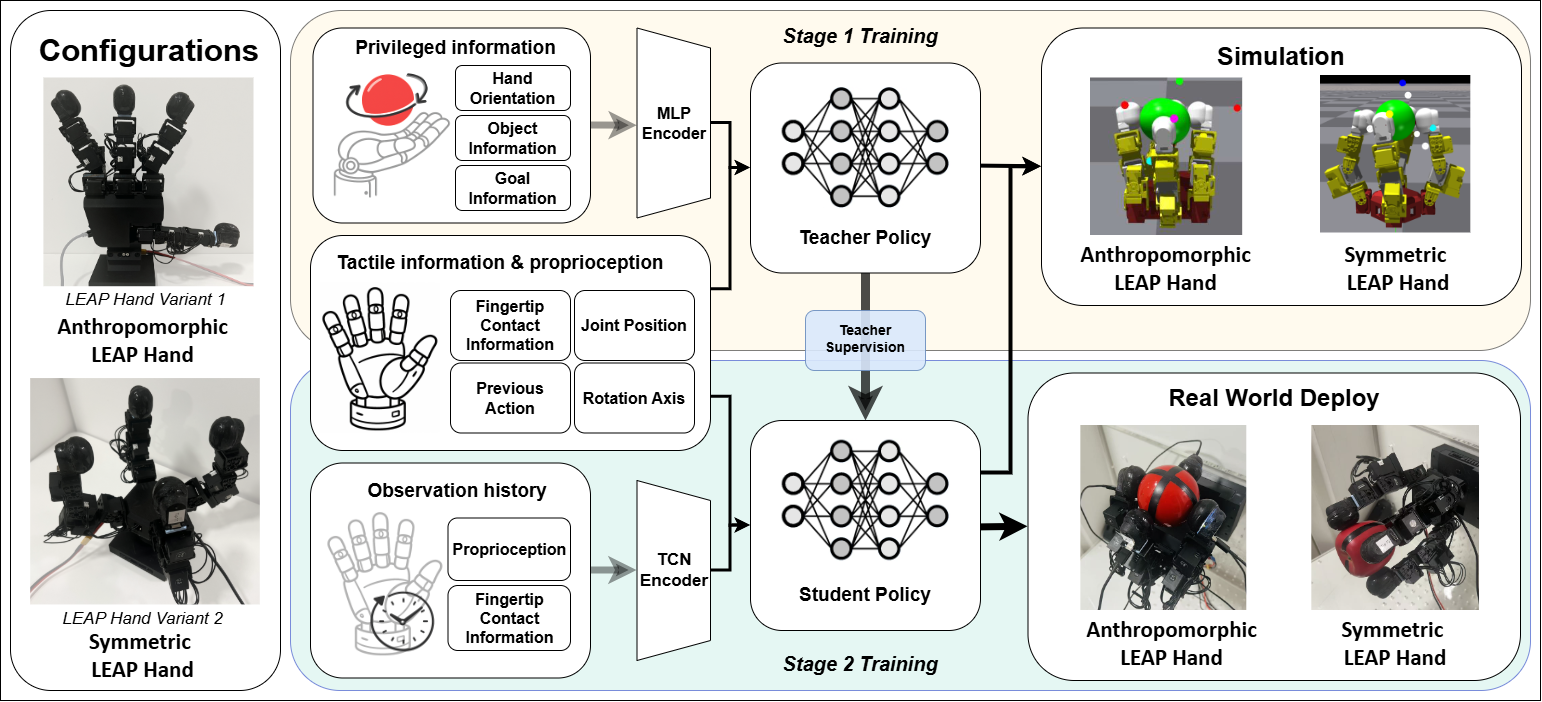}
    \caption{Overview of the approach. \textit{Left: }the two LEAP Hand embodiments. \textit{Right: }teacher–student training and deployment. The teacher is trained in simulation using deployable observations and privileged information. The student is trained under teacher supervision using the current deployable observation and a latent estimate of privileged information inferred from a finite history of deployable observations and commands. The trained student policy is deployed on the corresponding physical hand.}
    \label{fig:workflow}
\end{figure*}


\section{Related Work}
\label{sec:related_work}
\textbf{Tactile Feedback for Robotic Manipulation.}
Tactile manipulation spans explicit feedback control and learned policies. Li et al. use pressure-array features for tactile servoing~\cite{li2013servoing}, while van Hoof et al. learn object rolling with distributed tactile feedback on the compliant ReFlex hand~\cite{van2015learning}. On the torque-controlled DLR Hand-II, joint sensing supports cube manipulation~\cite{sievers2022Icra} and shape-conditioned reorientation of diverse objects~\cite{pitz2024learning}. Distributed binary contact on Allegro hands enables unseen-object rotation~\cite{yin2023rotating} and, in DexTouch, object seeking and manipulation without vision~\cite{lee2024dextouch}. Optical sensors such as the TacTip family infer local interaction from deformation of a compliant sensing surface~\cite{ward2018tactip,lepora2021soft}. Tactile Gym 2.0 compares TacTip, DIGIT, and DigiTac sensors under a common sim-to-real learning framework for object pushing, edge following, and surface following~\cite{lin2022tactilegym}. These systems differ in sensing coverage, information content, and robot mechanics.

\textbf{Learning Continuous In-Hand Rotation.}
Simulation-trained policies support dexterous reorientation on the Shadow Hand~\cite{andrychowicz2020learning} and across diverse object geometries~\cite{chen2022system}. For continuous rotation, rapid motor adaptation uses proprioceptive history~\cite{qi2023inhand}, whereas RotateIt combines vision, optical tactile information, and proprioception~\cite{qi2023general}. AnyRotate introduces gravity-invariant multi-axis rotation with contact-pose and force features, teacher--student training, and curricula~\cite{yang2024anyrotate}. It also evaluates tactile observation variants and analyzes reactive contact behavior on an Allegro hand. These observation choices distinguish information about fingertip
engagement from more detailed descriptions of contact geometry and
loading. We examine how the gains from these successive levels of
information vary with finger arrangement during sustained rotation.

\textbf{Hand Design and Comparative Evaluation.}
Comparative studies assess in-hand dexterity through manipulation tasks and measures of hand capability. The Elliott and Connolly Benchmark provides a common set of in-hand manipulation tasks~\cite{coulson2021benchmark}, while KaRMA compares feasible translation and reorientation through rolling contact in a two-finger pinch across 16 hand designs~\cite{peticco2026karma}. Complementing these evaluations, Almeida et al. investigate how tactile feedback from different finger and palm regions affects reorientation on a simulated Shadow Hand~\cite{almeida2025role}. Kinematic capability and executed manipulation provide complementary
views of hand design. Reachable motions describe the opportunities
afforded by geometry; contact persistence, contact changes, and
accumulated loading describe how the fingers participate as a task is
performed. This distinction motivates assessing workspace arrangement
together with measured interaction and task outcomes.

\section{Methods}\label{sec:exps_ressults}
\subsection{Task Definition}
\label{sec:Task_Definition}
We consider the AnyRotate task ~\cite{yang2024anyrotate}, which requires the hand to continuously rotate an object about a commanded rotation axis through fingertip manipulation under different hand orientations relative to gravity. Starting from an initial grasp, the policy sequentially tracks orientation targets separated by 30°. Target attainment is evaluated using six virtual keypoints located along the positive and negative directions of the three axes of the object coordinate frame. A subgoal is considered completed when the sum of Euclidean distances between the six corresponding object–target keypoint pairs does not exceed 0.015 m, thereby accounting for both position and orientation. Upon completion, the target advances without resetting the grasp. An episode terminates when the pose error becomes excessive, the rotation-axis deviation exceeds $45^\circ$, or all fingertip contacts are lost.

\subsection{Hand Embodiments}
To study how hand morphology affects tactile in-hand rotation performance, we compare two customized LEAP Hand embodiments equipped with TacTip
fingertips: an anthropomorphic hand and a symmetric hand (Fig.~\ref{fig:workflow}).
Both have four fingers and 16 actuated joints, but differ in finger
arrangement and fingertip orientation.

\subsubsection{Anthropomorphic LEAP Hand}

The Anthropomorphic LEAP Hand (top left of Fig.~\ref{fig:workflow}) is derived from the original CMU LEAP Hand~\cite{shaw2023leap}. It preserves the approximately anthropomorphic arrangement of three adjacent fingers and an opposing thumb, while adding TacTip tactile sensors to all fingertips. The fingertip sensing surfaces are oriented toward the central manipulation workspace: the index and little fingertips are yawed inward toward the middle finger, the middle fingertip remains aligned with the palm-forward direction, and the thumb fingertip is oriented between the thumb-forward and palm-forward directions. Together, these fingertip orientations define a partially enclosed manipulation region above the palm.

\subsubsection{Symmetric LEAP Hand}

The Symmetric LEAP Hand (bottom left of Fig.~\ref{fig:workflow}) is a task-oriented reconfiguration of the same LEAP platform. Its palm adopts an octagonal geometry, and the four fingers are mounted on four non-adjacent long edges so that opposing finger pairs are approximately orthogonal in the palm plane. Unlike the Anthropomorphic LEAP Hand, all fingertip sensing surfaces are aligned with the palm-forward direction. This arrangement distributes the four fingers more symmetrically
around the central manipulation region.

\subsection{Reinforcement Learning Structure}
\label{sec:RL}
\subsubsection{State, Observations, and Actions}

We model the underlying manipulation dynamics as a Markov decision process
\(\mathcal{M}=(\mathcal{S},\mathcal{A},P,r,\rho_0,\gamma)\), where \(\mathcal{S}\) and \(\mathcal{A}\) are the state and action spaces, \(P\) is the transition distribution, \(r\) is the reward, \(\rho_0\) is the initial-state distribution, and \(\gamma\) is the discount factor. The state includes hand and object dynamics, the active subgoal, controller memory, and physical parameters. The objective is to learn a policy \(\pi\) that maximizes
\begin{equation}
    J(\pi)
    =
    \mathbb{E}_{\rho_0,P,\pi}
    \left[
        \sum_{t=0}^{T-1}\gamma^t r_t
    \right],
\end{equation}
where \(T\) is the episode length.

The policy receives a partial observation \(\mathbf{o}_t\) containing joint positions, joint-position targets, the previous control increment, and the commanded rotation axis. Full tactile observations additionally include fingertip contact indicators, force magnitudes, and two-dimensional contact-pose features. Binary touch retains the contact indicators, while proprioception-only observations omit tactile channels. Task conditioning is retained in all conditions. Each observation ablation is applied consistently to both the current observation and the history provided to the student encoder. The policy outputs normalized joint-position increments. These increments are scaled, filtered, and bounded before updating the joint targets:

\begin{equation}
\begin{aligned}
    \mathbf{v}_t
    &=
    \eta\alpha\mathbf{a}_t
    +(1-\eta)\mathbf{v}_{t-1},
    \\
    \mathbf{q}^{\mathrm{cmd}}_{t+1}
    &=
    \mathbf{q}^{\mathrm{cmd}}_t
    +
    \operatorname{clip}
    \left(
        \mathbf{v}_t,
        -\mathbf{d}_{\max},
        \mathbf{d}_{\max}
    \right).
\end{aligned}
\end{equation}
Here, \(\alpha\) is the action scale, \(\eta\) is the filter coefficient, and \(\mathbf{d}_{\max}\) bounds each joint-target increment. The policy updates targets at 20~Hz, and joint PD controllers track the resulting commands.

\subsubsection{Reward Design}

We group the weighted reward components as
\begin{equation}
    r_t
    =
    r_t^{\mathrm{rot}}
    +r_t^{\mathrm{kp}}
    +\lambda_r r_t^{\mathrm{aux}}
    +r_t^{\mathrm{event}}.
\end{equation}
The rotational term \(r_t^{\mathrm{rot}}\) rewards clipped incremental progress about the commanded axis, while \(r_t^{\mathrm{kp}}\) assigns a bounded reward based on the distances between corresponding object and target keypoints, with smaller distances producing higher rewards. The auxiliary term \(r_t^{\mathrm{aux}}\) rewards multi-finger contact and penalizes non-fingertip contact, joint torque, mechanical power, and angular speed outside a prescribed range. The factor \(\lambda_r\in[0,1]\) controls the auxiliary reward curriculum. The event term \(r_t^{\mathrm{event}}\) provides a subgoal-completion bonus and an early-failure penalty; reaching the time limit incurs no failure penalty.

\subsubsection{Teacher--Student Learning and Curricula}
We use the teacher–student structure for policy training. The teacher combines policy observations with an eight-dimensional encoding of privileged simulator information. The MLP encoder and policy are jointly trained using proximal policy optimization (PPO)~\cite{schulman2017proximal}. Performance-based curricula scale gravity and auxiliary reward weights from zero to their full values as rotational performance improves.

The student inherits the teacher’s actor parameters and estimates the latent from a 30-step rolling history of observations and commands using a temporal convolutional encoder. The teacher remains frozen while the student actor and history encoder are jointly trained on student-generated rollouts using latent and action distillation with entropy regularization. Student training uses full gravity and auxiliary reward weights.

\subsubsection{Domain Randomization}
Object geometry and dimensions are sampled from the training object collection, while object mass, center-of-mass offset, and PD gains are randomized across parallel environments. At episode reset, initial hand--object states are sampled from cached grasps, hand orientation is randomized, and commanded rotation axes are sampled over the unit sphere. Random noise is added to joint-position and tactile contact-pose observations. Intermittent mass-scaled Gaussian forces with exponential decay further perturb the object during rollouts.

\subsection{Tactile Sensing and Perception}
\label{sec:Tactile Sensing and Perception}
We equip both hand embodiments with fingertip-mounted TacTip optical tactile sensors~\cite{lepora2021soft}. Their compliant skin carries internal pins whose marked tips are imaged by a camera. Contact deforms the skin and displaces the markers, producing tactile images that encode local mechanical interaction.

The contact representation and perception pipeline provide a common observation interface in simulation and on hardware. For fingertip \(i\in\{1,\ldots,4\}\) at time \(t\), the descriptor before masking and normalization is
\begin{equation}
    \mathbf{x}^{\mathrm{tac}}_{i,t}
    =
    \left[
        c_{i,t},\,
        \vartheta_{i,t},\,
        \varphi_{i,t},\,
        f_{i,t}
    \right]^{\top},
\end{equation}
where \(c_{i,t}\in\{0,1\}\) indicates contact, \(\vartheta_{i,t}\) and \(\varphi_{i,t}\) describe contact pose in spherical coordinates within the local sensor frame, and
\begin{equation}
    f_{i,t}
    =
    \left\lVert\mathbf{F}_{i,t}\right\rVert_2
\end{equation}
is the magnitude of the three-dimensional contact force. These features describe fingertip engagement, contact geometry, and loading during rotation.

In simulation, contact positions and forces are obtained at the fingertip sensing surfaces and converted into the corresponding features. Force signals are temporally filtered, while contact-pose and force values are saturated and rescaled to align their ranges with those of the physical sensors.

For physical deployment, a tactile observation model is trained offline using images paired with contact-pose and force labels collected with a robot arm and a force/torque sensor. At runtime, the model estimates contact pose and force from each fingertip image. On hardware, contact is detected when the structural similarity index (SSIM)~\cite{wang2004ssim}  exceeds 0.5. The resulting binary contact estimate masks the pose and force predictions when no contact is detected. The resulting tactile features are combined with proprioceptive observations, providing both hand embodiments with the same types of local contact information.

\section{Experimental Setup}
\begin{figure}[t]
    \centering
    \includegraphics[width=\columnwidth]{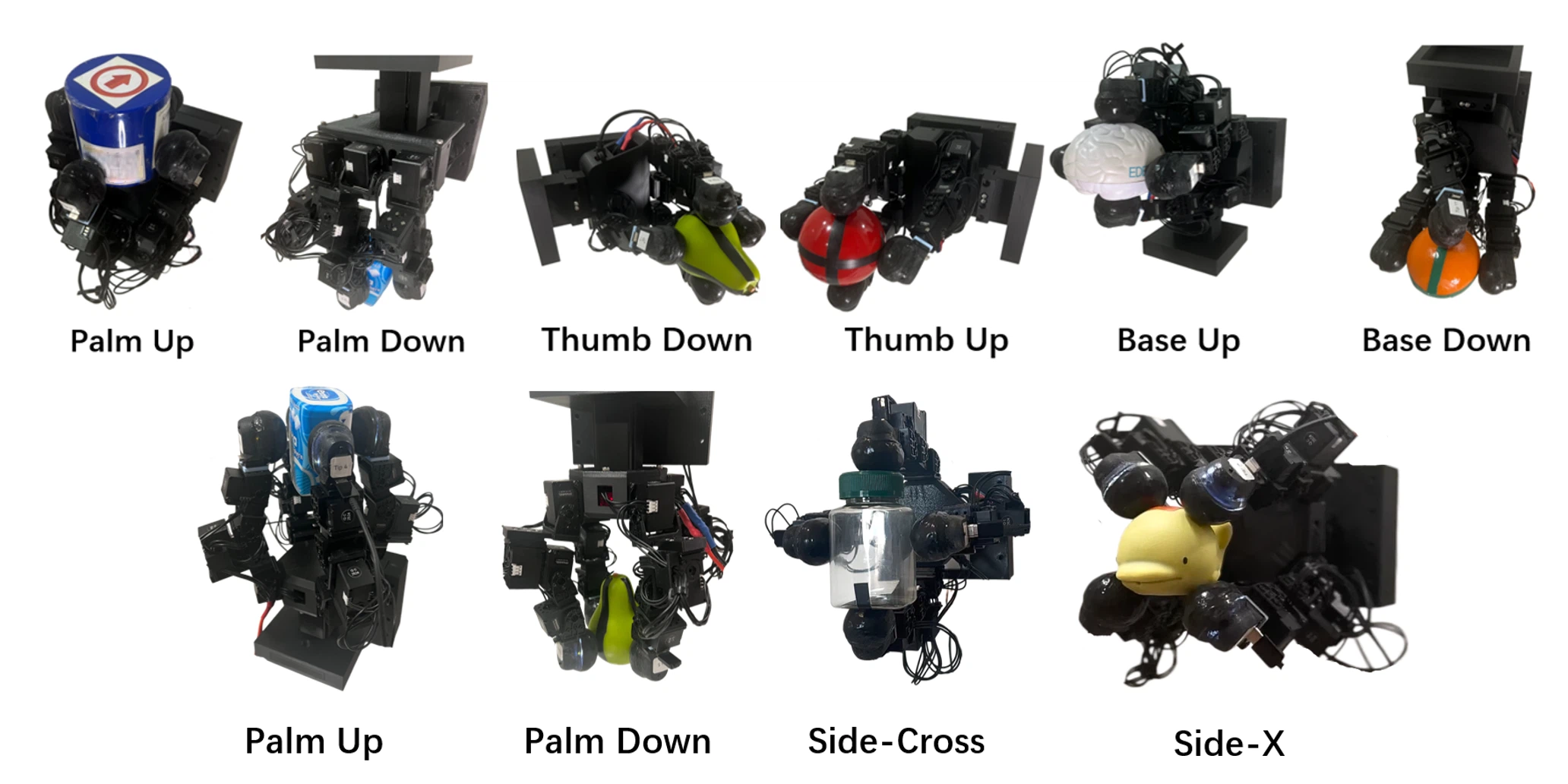}
    \caption{Hand orientations used in real-world evaluation for the Anthropomorphic and Symmetric LEAP Hand embodiments.}
    \label{fig:leapori}
\end{figure}

\subsection{Simulation Training and Experiments}

We train teacher and student policies in Isaac Gym~\cite{makoviychuk2021isaacgym} with a simulation
timestep of $1/60$~s and a policy control frequency of 20~Hz. The
training objects comprise cuboids and ellipsoids of varying sizes, and domain randomization is applied following Sec.~\ref{sec:RL}.

To reduce the sim-to-real discrepancy in hand dynamics, we perform
trajectory-based system identification. The identification recordings include joint trajectories collected
under multiple hand orientations relative to gravity. For the symmetric
hand, they include configurations with each of the four fingers pointing
downward, covering the corresponding changes in gravitational loading. Recorded joint commands are
replayed in simulation, and the covariance matrix adaptation evolution
strategy (CMA-ES)~\cite{hansen2001cmaes} optimizes PD gains, link masses, joint friction, and armature parameters to minimize the discrepancy between simulated and recorded joint trajectories. Simulation evaluation uses 12 held-out object models excluded
from training (Fig.~\ref{fig:simrealobjsetimg}). For each hand and observation
condition---full tactile, binary touch, or proprioception
only---the trained student policy is evaluated in 64 parallel
environments with randomly sampled objects, hand orientations,
and commanded rotation axes. Reported means are cumulative averages over completed episodes
lasting more than 20 control steps, pooled across the 64 parallel
environments. We report the values after approximately 200
evaluation iterations, when the monitored means have stabilized. Episodes last up to 600 control steps
(30~s), subject to the termination criteria in Sec.~\ref{sec:Task_Definition}.

We report mean episode duration and mean normalized rotation per
episode. Normalized rotation is accumulated as
\[
R_{\mathrm{rot}} = \frac{1}{3.14}\sum_t \Delta\theta_t,
\]
where $\Delta\theta_t$ is the signed angular increment in radians,
estimated by projecting the Euler-angle increments of the relative
object orientation onto the commanded rotation axis.
Duration is the elapsed time until termination or the evaluation
horizon.

\subsection{Real-World Experiments}
\label{sec:hardware_setup}

\begin{figure}[t]
    \centering
    \includegraphics[width=\columnwidth]{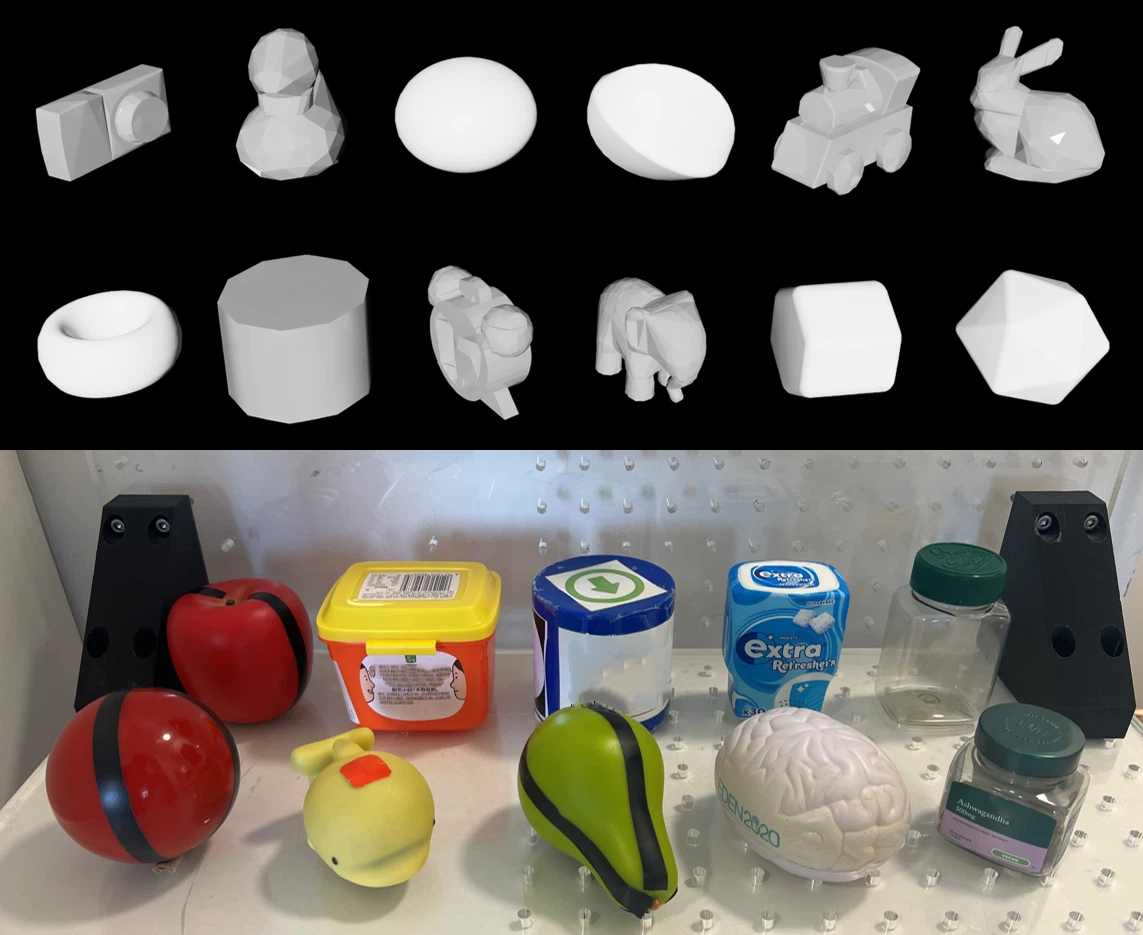}
    \caption{Top: Simulated object set. Bottom: real-world object set.}
    \label{fig:simrealobjsetimg}
\end{figure}

Both hands are mounted on a common fixture supporting
repeatable orientations relative to gravity (Fig.~\ref{fig:leapori}) and use the tactile perception pipeline described in Sec.~\ref{sec:Tactile Sensing and Perception}.
Full tactile policies are evaluated on ten physical objects
spanning different shapes, sizes, aspect ratios, and masses
(Fig.~\ref{fig:simrealobjsetimg}), with five trials per object and hand orientation.
Trials command rotation about the palm-aligned axis and last
up to 30~s.

Rotation count is estimated visually in revolutions.
Manipulation duration is the time until the object is dropped,
becomes stuck, or the rollout otherwise becomes invalid;
trials reaching the horizon are assigned 30~s. Results are
reported as mean and standard deviation. Overall rotation
count and duration are calculated by equally weighting the
orientation-group means, and overall rotation rate is their
ratio.

\subsection{Contact and Loading Metrics}
\label{sec:contact_loading_metrics}

Force recordings follow the object, hand-orientation, and
rotation-axis settings of the corresponding evaluation
protocols. Contact-time ratios and transition rates are
computed by dividing pooled contact durations and event
counts, respectively, by total recorded duration. Transitions
are changes in either direction of a finger's binary contact
state; resets are excluded. The multi-finger contact fraction
is the proportion of recorded time with at least three
fingertips in contact.

Let $f_i(t)$ denote the magnitude of fingertip $i$'s body net
contact force in simulation or its tactile-estimated force on
hardware. Simulator forces aggregate contacts on each
fingertip body. Total fingertip load $L(t)$ and each finger's
cumulative load share $s_i$ are
\begin{equation}
    L(t)=\sum_{i=1}^{4}f_i(t),
    \qquad
    s_i=\frac{\sum_t f_i(t)\Delta t}
                 {\sum_{j=1}^{4}\sum_t f_j(t)\Delta t},
\end{equation}
where $\Delta t$ is the sampling interval. Total load sums
force magnitudes rather than force vectors. The cumulative-load
coefficient of variation (CV) is the population standard
deviation of the four integrated fingertip loads divided by
their mean.

Hardware statistics for total fingertip load are computed from pooled
recorded samples, including samples outside detected contact. The
standard deviation (SD) describes variation across these samples, and
the upper load level is the value not exceeded by 95\% of them.
Mean force during contact is calculated separately for each finger
using only samples in which that finger is detected to be in contact.
For both simulation and hardware, cumulative load shares and their
CV are computed after pooling each finger's force-time integrals
across the corresponding recordings.

\section{Results}
\begin{table}[t]
\centering
\caption{Simulation performance under different observation conditions.}
\label{tab:sim_performance}
\small
\setlength{\tabcolsep}{3pt}
\begin{tabular*}{\columnwidth}{@{\extracolsep{\fill}}lccc@{}}
\toprule
Metric & \shortstack{Full\\tactile}
       & \shortstack{Binary\\touch}
       & \shortstack{Proprioception\\only} \\
\midrule
\multicolumn{4}{c}{\textbf{Anthropomorphic LEAP Hand}} \\
{Mean normalized rotation $\uparrow$}
    & 1.893 & 1.657 & 0.972 \\
Mean duration (s) $\uparrow$          & 26.83 & 24.49 & 21.47 \\
\midrule
\multicolumn{4}{c}{\textbf{Symmetric LEAP Hand}} \\
{Mean normalized rotation $\uparrow$}
    & 1.974 & 1.731 & 1.183 \\
Mean duration (s) $\uparrow$           & 25.73 & 23.05 & 21.92 \\
\bottomrule
\end{tabular*}
\begin{minipage}{\columnwidth}
\vspace{2pt}
\footnotesize
Normalized rotation is dimensionless and is computed by accumulating
signed angular increments along the commanded axis, divided by
3.14~rad.
\end{minipage}
\end{table}

\subsection{Simulation Results}
\label{sec:simulation_results}

For both hands, mean normalized rotation and episode duration
increase from proprioception-only observations to binary touch
and then to full tactile observations
(Table~\ref{tab:sim_performance}). The symmetric hand achieves
higher normalized rotation under every observation condition.
With full tactile observations, the anthropomorphic and symmetric
hands achieve mean normalized rotation values of 1.89 and 1.97,
respectively, with mean durations of 26.8 and 25.7~s.

\subsection{Real-World Results}
Table~\ref{tab:orientation_performance} reports the rotation count and manipulation duration for each orientation. For the symmetric hand, Side-Cross and Side-X denote side orientations with a finger or a gap between adjacent fingers uppermost, respectively. Each group comprises four orientations separated by 90° about the palm normal. We evaluate all four orientations in each group and average their mean rotation counts and manipulation durations to obtain the reported group means. 

\begin{figure}[t]
    \centering
    \includegraphics[width=\columnwidth]{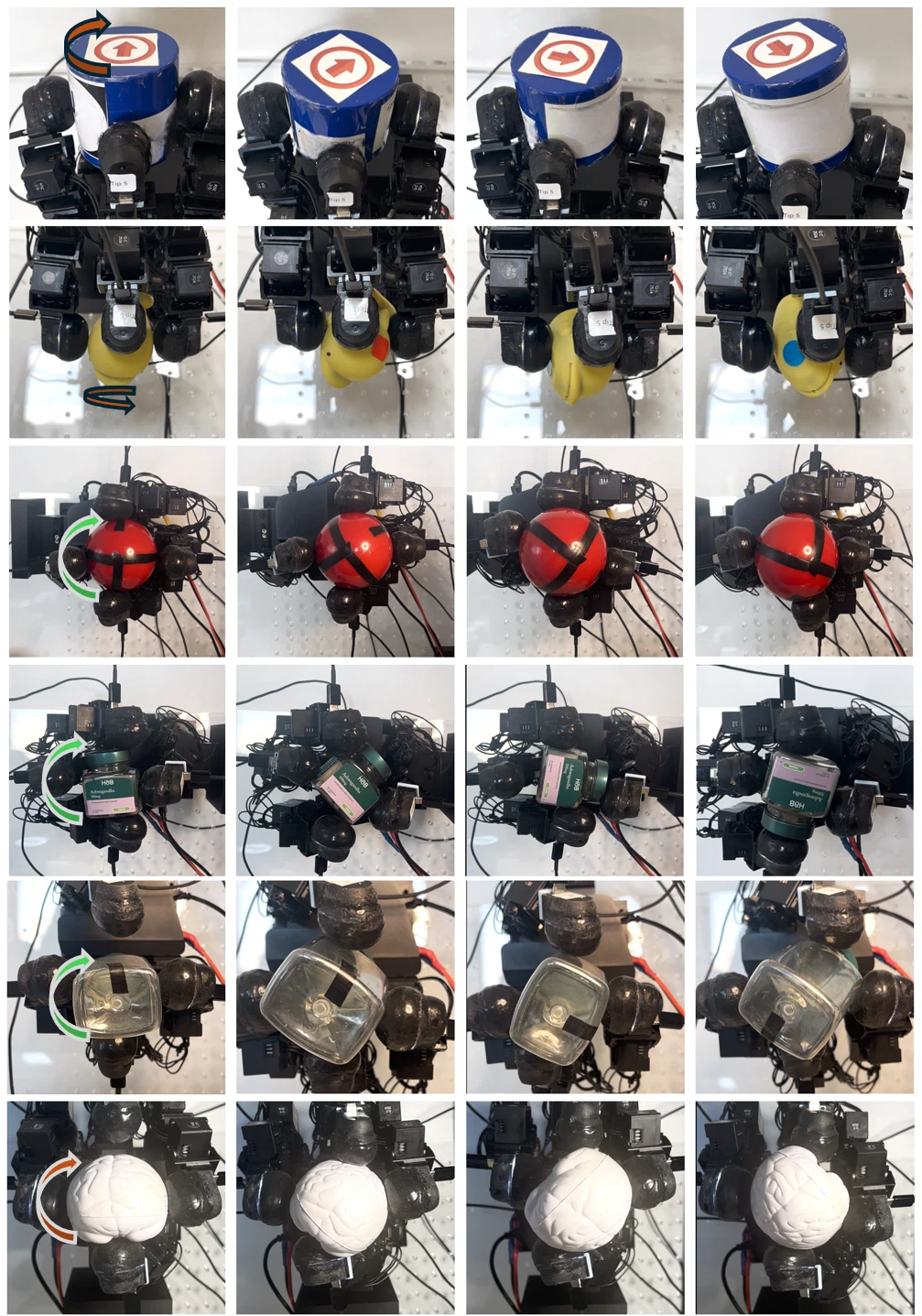}
    \caption{In-hand manipulation with the Anthropomorphic LEAP
Hand under different hand orientations.}
    \label{fig:leaphandreal}
\end{figure}

\begin{figure}[t]
    \centering
    \includegraphics[width=\columnwidth]{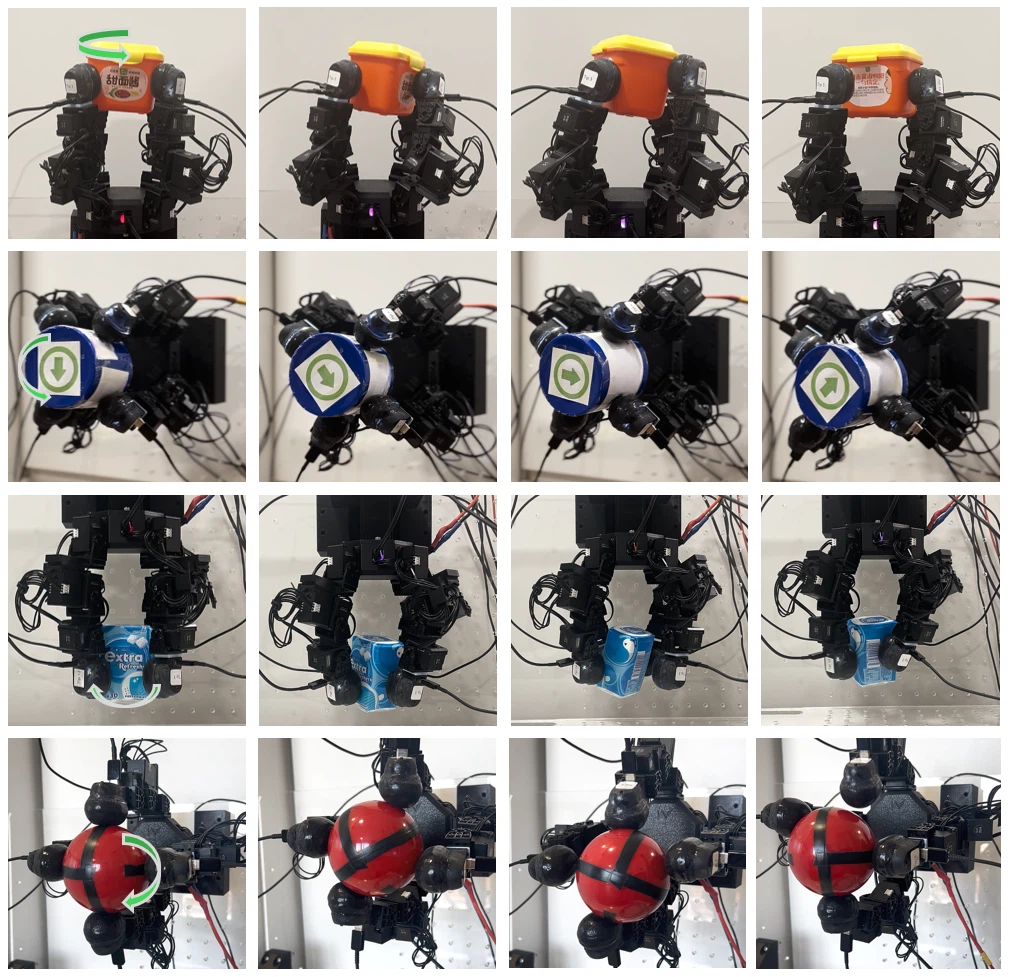}
    \caption{In-hand manipulation with the Symmetric LEAP Hand
under different hand orientations.}
    \label{fig:leaprotaterealS}
\end{figure}

Representative hardware rollouts are shown in Fig.~\ref{fig:leaphandreal} and Fig.~\ref{fig:leaprotaterealS}. In the palm-up and palm-down orientations shared by both embodiments, the symmetric hand achieves higher mean rotation counts and longer mean manipulation durations. In the palm-up condition, it completes 1.48 revolutions over 30.0~s, compared with 1.20 revolutions over 28.4~s for the anthropomorphic hand. In the palm-down condition, the corresponding values are 1.32 revolutions over 27.2~s and 1.01 revolutions over 25.9~s.

Across all tested orientations, the anthropomorphic hand achieves mean rotation counts ranging from 0.79 to 1.20 revolutions and mean durations ranging from 22.9 to 30.0~s. The symmetric hand achieves 1.30 revolutions over 28.9~s in Side-Cross and 1.05 revolutions over 26.0~s in Side-X. Its mean rotation count remains above one revolution in all four reported orientation groups.
\begin{table}[t]
    \centering
    \caption{Real-world manipulation performance across hand orientations.}
    \label{tab:orientation_performance}

    \footnotesize
    \setlength{\tabcolsep}{2pt}
    \renewcommand{\arraystretch}{1.08}

    \begin{tabularx}{\columnwidth}{
        @{}
        >{\centering\arraybackslash}p{0.32\columnwidth}
        *{2}{>{\centering\arraybackslash}X}
        @{}
    }
        \toprule
        Hand orientation
        & \shortstack{Rotation count\\(revolutions) $\uparrow$}
        & \shortstack{Manipulation\\duration (s) $\uparrow$} \\
        \midrule

        \multicolumn{3}{c}{\textbf{Anthropomorphic LEAP Hand}} \\
        \addlinespace[1pt]

        Palm Up
        & \(1.20 \pm 0.20\)
        & \(28.38 \pm 3.03\) \\
        \lightrowrule

        Palm Down
        & \(1.01 \pm 0.29\)
        & \(25.88 \pm 6.14\) \\
        \lightrowrule

        Thumb Up
        & \(0.94 \pm 0.56\)
        & \(30.00 \pm 0.00\) \\
        \lightrowrule

        Thumb Down
        & \(0.88 \pm 0.22\)
        & \(27.40 \pm 4.92\) \\
        \lightrowrule

        Base Up
        & \(1.04 \pm 0.30\)
        & \(22.92 \pm 7.08\) \\
        \lightrowrule

        Base Down
        & \(0.79 \pm 0.25\)
        & \(25.67 \pm 6.71\) \\

        \addlinespace[2pt]
        \midrule
        \addlinespace[2pt]

        \multicolumn{3}{c}{\textbf{Symmetric LEAP Hand}} \\
        \addlinespace[1pt]

        Palm Up
        & \(1.48 \pm 0.28\)
        & \(30.00 \pm 0.00\) \\
        \lightrowrule

        Palm Down
        & \(1.32 \pm 0.24\)
        & \(27.19 \pm 4.14\) \\
        \lightrowrule

        Side-Cross
        & \(1.30 \pm 0.42\)
        & \(28.91 \pm 3.02\) \\
        \lightrowrule

        Side-X
        & \(1.05 \pm 0.29\)
        & \(26.04 \pm 4.36\) \\

        \bottomrule
    \end{tabularx}
\end{table}

\section{Analysis}
\label{sec:analysis}
\begin{figure}[t]
    \centering
    \includegraphics[width=\columnwidth]{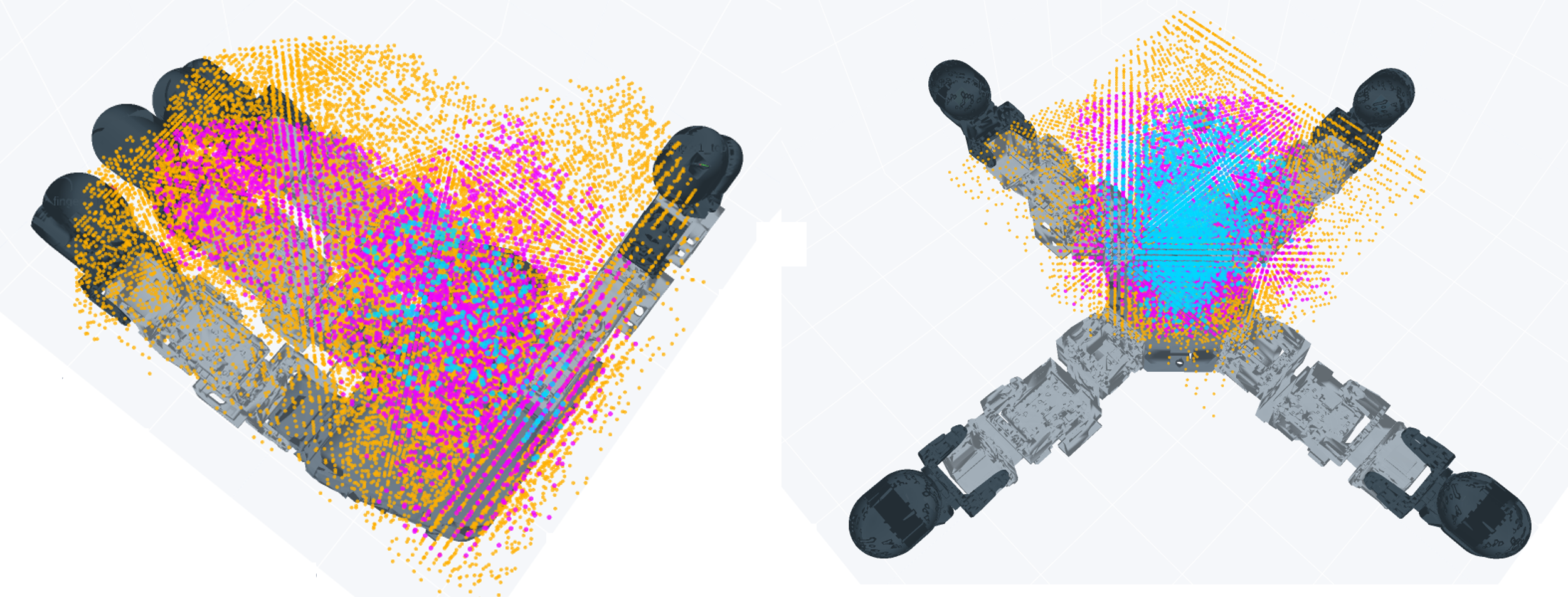}
    \caption{Fingertip-centre workspaces for the anthropomorphic (left)
and symmetric (right) hands. Orange, red, and blue denote cells
reachable by exactly two, three, and four fingertips, respectively,
across sampled configurations. Of 40,000 sampled points, 8,000 are displayed.}
    \label{fig:handworkspace}
\end{figure}

We relate workspace arrangement to finger participation and
load sharing, and examine how tactile information changes
task progress and fingertip loading.

\begin{table}[t]
\centering
\caption{Hardware finger participation and fingertip loading.}
\label{tab:hardware_participation_load}
\small
\setlength{\tabcolsep}{2.5pt}
\begin{tabular*}{\columnwidth}{@{\extracolsep{\fill}}lcccc@{}}
\toprule
Metric & \shortstack{Thumb/\\Finger 1}
       & \shortstack{Index/\\Finger 2}
       & \shortstack{Middle/\\Finger 3}
       & \shortstack{Little/\\Finger 4} \\
\midrule
\multicolumn{5}{c}{\textbf{Anthropomorphic LEAP Hand}} \\
Contact time (\%) & 91.23 & 81.62 & 93.12 & 75.98 \\
Load share (\%)   & 28.99 & 22.40 & 28.24 & 20.37 \\
Transition rate (s$^{-1}$)
    & 1.101 & 2.454 & 0.875 & 2.261 \\
\shortstack[l]{Mean force\\during contact (N)}
    & 4.13 & 3.57 & 3.94 & 3.48 \\
\midrule
\multicolumn{5}{c}{\textbf{Symmetric LEAP Hand}} \\
Contact time (\%) & 85.45 & 84.23 & 82.94 & 85.61 \\
Load share (\%)   & 25.82 & 23.16 & 26.47 & 24.55 \\
Transition rate (s$^{-1}$)
    & 1.603 & 1.758 & 1.642 & 1.597 \\
\shortstack[l]{Mean force\\during contact (N)}
    & 3.15 & 2.86 & 3.32 & 2.99 \\
\bottomrule
\end{tabular*}
\vspace{2pt}
\begin{minipage}{\columnwidth}
\vspace{2pt}
\footnotesize
Names refer to anthropomorphic fingers; numbers identify
symmetric-hand channels. Transition rates use the full-recording
event counts. Metric definitions are given in
Sec.~\ref{sec:contact_loading_metrics}.
\end{minipage}
\end{table}

\subsection{Workspace Arrangement and Finger Participation}
Fig.~\ref{fig:handworkspace} shows fingertip-centre workspaces
obtained by sampling joint configurations within their limits
and applying forward kinematics. Overlap indicates geometric
reachability, not simultaneous contact feasibility. The
anthropomorphic arrangement provides unequal approach
directions through three adjacent fingers and an opposing
thumb; the symmetric layout places approximately orthogonal
opposing pairs around the manipulation region.

The hardware recordings reveal corresponding differences in finger
participation (Table~\ref{tab:hardware_participation_load}). The
anthropomorphic thumb and middle finger have the most persistent contact
and the highest mean forces during detected contact. Together, they
account for 57.2\% of cumulative fingertip load and approximately
29.5\% of contact-state transitions. The index and little fingers
contribute the remaining 42.8\% of load and 70.5\% of transitions.
Persistent loading and contact changes are therefore concentrated in
different finger pairs within this hand.

For the symmetric hand, contact-time ratios span 82.9\%--85.6\%,
transition rates span 1.60--1.76~s$^{-1}$, and load shares span
23.3\%--26.5\%. Its four fingers thus show more evenly distributed
participation across these measures. The total transition rates remain
close, at 6.69 and 6.60~s$^{-1}$ for the anthropomorphic and symmetric
hands, respectively. The difference between layouts concerns how contact
changes and accumulated loading are distributed across fingers at
comparable overall rates of contact-state change.

The symmetric hand also has lower cumulative-load CV under each
simulated observation condition (Table~\ref{tab:sim_fingertip_load}).
The pooled hardware CV values are 0.148 for the anthropomorphic hand
and 0.051 for the symmetric hand (Table~\ref{tab:hardware_total_load}).
This correspondence concerns the distribution of accumulated loading
across fingers in the two evaluation domains.

\begin{table}[t]
\centering
\caption{Fingertip loading and cumulative load sharing in simulation.}
\label{tab:sim_fingertip_load}
\small
\setlength{\tabcolsep}{3pt}
\begin{tabular*}{\columnwidth}{@{\extracolsep{\fill}}lccc@{}}
\toprule
Metric & \shortstack{Full\\tactile}
       & \shortstack{Binary\\touch}
       & \shortstack{Proprioception\\only} \\
\midrule
\multicolumn{4}{c}{\textbf{Anthropomorphic LEAP Hand}} \\
Mean total load (N) & 10.31 & 12.28 & 13.49 \\
Cumulative-load CV & 0.138 & 0.135 & 0.085 \\
\midrule
\multicolumn{4}{c}{\textbf{Symmetric LEAP Hand}} \\
Mean total load (N) & 11.63 & 11.93 & 13.01 \\
Cumulative-load CV & 0.050 & 0.024 & 0.029 \\
\bottomrule
\end{tabular*}
\vspace{2pt}
\begin{minipage}{\columnwidth}
\vspace{2pt}
\footnotesize
CV is dimensionless and measures variation across cumulative
fingertip loads; lower values indicate more uniform cumulative
load sharing.
\end{minipage}
\end{table}

\begin{table}[t]
\centering
\caption{Hardware fingertip loading and cumulative load sharing.}
\label{tab:hardware_total_load}
\small
\setlength{\tabcolsep}{3pt}
\begin{tabular*}{\columnwidth}{@{\extracolsep{\fill}}lcc@{}}
\toprule
Metric & \shortstack{Anthropomorphic\\LEAP Hand}
       & \shortstack{Symmetric\\LEAP Hand} \\
\midrule
Mean load $\pm$ SD (N) & $13.01 \pm 5.09$ & $10.42 \pm 3.37$ \\
Upper load level (N) & 20.86 & 16.44 \\
Cumulative-load CV & 0.148 & 0.051 \\
\bottomrule
\end{tabular*}
\vspace{2pt}
\begin{minipage}{\columnwidth}
\vspace{2pt}
\footnotesize
Load statistics use pooled force samples. Cumulative-load CV
is computed from the four fingertip load integrals pooled
across recordings. Upper load level is defined in
Sec.~\ref{sec:contact_loading_metrics}.
\end{minipage}
\end{table}

\subsection{Tactile Benefits and Fingertip Loading}
Full tactile observations increase mean normalized rotation over
proprioception alone by 94.8\% for the anthropomorphic hand and
66.9\% for the symmetric hand (Table~\ref{tab:sim_performance}). Binary touch increases
this metric by 0.685 and 0.548, respectively, relative to
proprioception alone. Jointly adding contact-pose and force features
yields further increases of 0.236 and 0.243, corresponding to
14.2\% and 14.0\% over binary touch. These richer features therefore
provide similar additional gains across the two layouts.

These observation-dependent gains also change the numerical gap
between the hands. The symmetric hand's mean normalized rotation
exceeds that of the anthropomorphic hand by 0.211 under
proprioception alone, compared with 0.074 under binary touch and
0.081 under full tactile observations. Thus, the two tactile
conditions yield closer mean normalized rotation values across
layouts. Most of this reduction occurs when contact-state
information becomes available, corresponding to the larger
binary-touch gain of the anthropomorphic hand.

Mean total fingertip load decreases across the same observation
levels (Table~\ref{tab:sim_fingertip_load}). From proprioception
alone to full tactile observations, it falls from 13.5 to
10.3~N for the anthropomorphic hand and from 13.0 to
11.6~N for the symmetric hand, reductions of 23.6\% and
10.6\%, respectively. In both embodiments, higher mean normalized rotation and longer manipulation accompany lower average fingertip loading.

Comparing full tactile observations directly with binary touch reveals
a further difference between layouts. The additional rotation gains
are similar, while mean total load decreases by 16.0\% for the
anthropomorphic hand and 2.5\% for the symmetric hand. Richer
observations therefore accompany comparable improvements in rotational progress and different changes in average loading. 

\subsection{Hardware Loading and Manipulation Outcomes}

The additional hardware recordings show a lower mean total
load for the symmetric hand: 10.4~N versus 13.0~N,
a reduction of 19.9\%
(Table~\ref{tab:hardware_total_load}). Its upper load level
is also lower, at 16.4~N versus 20.9~N. Together with the
cumulative-load CV, these measurements show lower total
loading and more uniform load sharing in the symmetric-hand
recordings. The loading difference also appears when force is evaluated only during
detected contact. Mean forces range from 3.48 to 4.13~N across the
anthropomorphic fingers and from 2.86 to 3.32~N across the symmetric
fingers (Table~\ref{tab:hardware_participation_load}). Thus, the lower
aggregate loading of the symmetric hand is accompanied by lower
per-finger force levels during contact. 

In the hardware performance evaluation, the symmetric hand
achieves a 32.0\% higher mean rotation count and a 25.8\%
higher aggregate rotation rate, together with longer mean
manipulation duration. The anthropomorphic and symmetric
hands maintain at least three fingertip contacts for 95.4\%
and 95.5\% of the recorded time, respectively. The performance
difference therefore occurs with similarly persistent
multi-finger engagement.

Simulation samples arbitrary rotation axes, while hardware
trials use palm-axis rotation. Differences between compliant
physical fingertips and rigid simulated fingertips, together
with residual joint-motion mismatch after CMA-ES
identification, may also contribute to differences in loading
and duration between domains.

\subsection{Implications for Hand Design and Tactile Control}

These findings suggest an evaluation sequence for hand design.
Workspace maps first describe the regions reachable by the fingertips.
Finger-level recordings then identify how the controller distributes
persistent contact, contact changes, and loading within that workspace.
For example, a modification to finger placement can be assessed by
whether it alters these distributions and whether the altered interaction
accompanies greater rotational progress or longer manipulation. This
connects a geometric design change to observable behavior and a task
outcome.

Selection of tactile inputs can be evaluated in the same way. For the
evaluated task, the results support retaining contact-state information
and show an additional benefit from the combined contact-pose and force
features for both layouts. Similar gains in normalized rotation can accompany different reductions in mean load. Comparing progress, duration, and loading together provides a basis for assessing
what a sensing choice contributes on each hand.

\section{Conclusion and Future Work}
This study compared two LEAP-based finger arrangements using
a common tactile interface and teacher--student learning framework. In simulation, binary touch improves mean normalized rotation and manipulation duration over proprioception alone, and full tactile
observations provide further gains while reducing mean fingertip
loading. The additional gains in mean normalized rotation from contact-pose and force features are similar across the two layouts, despite their
different patterns of finger participation.

Hardware contact measurements reveal more persistent
contributions from the anthropomorphic thumb and middle
finger, with more frequent contact changes at the index and
little fingers. The symmetric hand shows more evenly
distributed contact participation on hardware and cumulative
load sharing in both domains. It also achieves a higher
rotation count and rotation rate, together with longer
manipulation duration, in the hardware performance evaluation.
Additional hardware force recordings show lower mean total
fingertip loading for this layout. These findings motivate
evaluating geometric reachability together with observed
finger contributions, achieved task progress, and loading
during manipulation.

Future work will extend the comparison to additional hand
layouts and tasks, evaluate reduced tactile observations on
hardware, and combine object-pose tracking with contact-resolved
force measurements to examine how contact renewal contributes
to rotational motion.

\bibliographystyle{IEEEtran}
\bibliography{reference}
\end{document}